\documentclass[11pt]{article}

\usepackage[final]{acl}

\usepackage{ulem}

\usepackage{times}
\usepackage{makecell}
\usepackage{multirow}
\usepackage{latexsym}
\usepackage{natbib}
\usepackage{amsmath,amssymb,amsfonts}
\usepackage{graphicx}
\usepackage{subcaption}
\usepackage{textcomp}
\usepackage{xcolor}
\usepackage{url}
\usepackage{algorithm}
\usepackage{algpseudocode}
\usepackage{tabularx}
\usepackage{adjustbox}
\usepackage{booktabs}
\usepackage{tikz}
\usetikzlibrary{shapes,arrows}
\usepackage{circuitikz}
\usepackage{enumitem}
\usepackage{wasysym}
\usepackage{colortbl}
\usepackage[most]{tcolorbox}
\usepackage[table]{xcolor}
\definecolor{cellbg}{RGB}{230, 245, 255} 
\definecolor{headerbg}{RGB}{230, 245, 255} 
\definecolor{red4}{HTML}{febf92}
\definecolor{dt}{gray}{0.7}
\def\BibTeX{{\rm B\kern-.05em{\sc i\kern-.025em b}\kern-.08em
    T\kern-.1667em\lower.7ex\hbox{E}\kern-.125emX}}

\usepackage[T1]{fontenc}
\usepackage[utf8]{inputenc}

\usepackage{microtype}

\usepackage{inconsolata}

\usepackage{graphicx}
\usepackage{hyperref}

\title{TIR-Flow: Active Video Search and Reasoning with Frozen VLMs}

\author{
 \textbf{Hongbo Jin\thanks{Equal contribution}} \quad
 \textbf{Siyi Xie\footnotemark[1]} \quad
 \textbf{Jiayu Ding\footnotemark[1]} \quad
 \textbf{Kuanwei Lin} \quad
 \textbf{Ge Li\thanks{Corresponding author}}
\\
 School of Electronic and Computer Engineering,\\
 Peking University
\\
 \small{
   \textbf{Correspondence:} \href{hbjin25@stu.pku.edu.cn}{hbjin25@stu.pku.edu.cn}
 }
}

\begin{document}
\maketitle
\begin{abstract}
While Large Video-Language Models (Video-LLMs) have achieved remarkable progress in perception, their reasoning capabilities remain a bottleneck. Existing solutions typically resort to a heavy "data engineering" paradigm—synthesizing large-scale Chain-of-Thought (CoT) datasets followed by Supervised Fine-Tuning (SFT) and Reinforcement Learning (RL).
This pipeline primarily optimizes probability sampling efficiency and aligns output distributions,
but fails to activate the intrinsic intelligence required for dynamic visual exploration.
In this work, we propose \textbf{TIR-Flow},
a novel framework that shifts the paradigm from passive processing to active video searching and reasoning without additional data or parameter updating.
Concretely, our framework operates through three synergistic modules: \textbf{HDD } decomposes complex queries into a set of verifiable sub-tasks; \textbf{HAP} actively directs visual attention to gather high-resolution evidence for hypothesis validation;
\textbf{EBA} maintains a persistent workspace to accumulate and update the discovered clues for logical reasoning.
Extensive experiments on seven benchmarks demonstrate that TIR-Flow significantly outperforms recent strong baselines, delivering an average performance boost of \textbf{5.9\%}, with gains reaching \textbf{10.5\%} on Egoschema.
Our analysis confirms that empowering frozen VLMs with System-2-like active perception is a scalable path toward solving long-horizon video reasoning.

\end{abstract}

\section{Introduction}
\label{sec:intro}
The rapid evolution of Large Vision-Language Models (VLMs) has revolutionized the field of video understanding~\cite{zohar2025apollo, zhang2025videollama, li2024llava,xu2025slowfastllava15familytokenefficientvideo}.
While these models demonstrate impressive capabilities in basic visual perception, they fundamentally struggle with complex reasoning over long horizons.
% These models have demonstrated impressive capabilities in describing visual content and answering fact-based questions.
% However,
% complex reasoning over long videos remains a persistent bottleneck.
When faced with tasks requiring causal deduction or fine discrimination, standard Video-LLMs often miss critical details.

To bridge this gap, prevailing research relies heavily on data-centric optimization. Approaches such as synthesizing Video Chain-of-Thought (CoT) datasets~\cite{qi2025vcrbenchcomprehensiveevaluationframework, zhang2025videocotcomprehensivedatasetspatiotemporal} or applying Reinforcement Learning (RL)~\cite{feng2025videor1reinforcingvideoreasoning,li2025videochat,wang-2025-open-r1-video} aim to refine the reasoning distribution.

However, this paradigm faces a critical limitation: it relies heavily on extensive data engineering to align the model with specific reasoning patterns. Consequently, while the model achieves high scores on familiar tasks, it struggles to generalize to out-of-distribution (OOD) scenarios characterized by unseen visual complexities.
Drawing inspiration from human cognitive processes~\cite{evans2003two},
we posit that complex video reasoning requires an active search mechanism. 
Just as a human expert does not passively watch a video but actively rewinds, zooms in,
and verifies specific hypotheses, a VLM should be empowered to "plan, look, and verify" before answering.

\begin{figure*}[h]
    \centering
    \includegraphics[width=\linewidth, trim=25pt 120pt 15pt 5pt, clip]{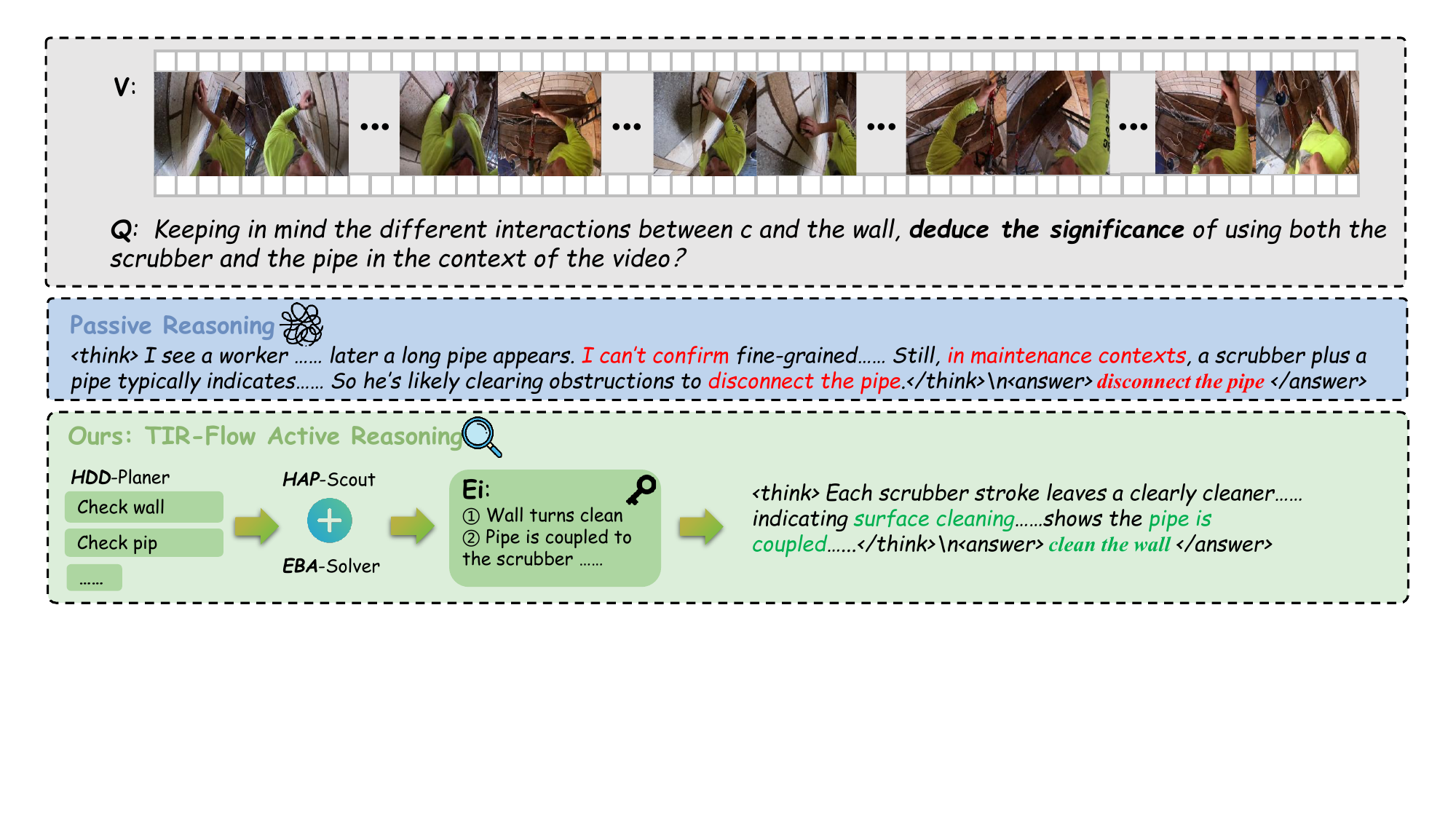}
    \caption{\textbf{Comparison of Reasoning Trajectories between Passive CoT and TIR-Flow Paradigm.} Without an explicit framework to guide the model in extracting and verifying visual information, it is challenging for the system to perform logical deductions grounded in the intrinsic content of the video.}
    \label{fig:example}
\end{figure*}

To this end, we introduce TIR-Flow (Temporal Intervention Reasoning Flow),
a unified, training-free framework designed to unlock the latent reasoning potential of foundation models.
TIR-Flow shifts the focus from static parameter optimization to  active perception.
This framework utilizes a cascaded modular architecture to dynamically bridge the gap between low resolution global contexts and high fidelity local evidence.
As illustrated in~\autoref{fig:example}, 
The framework operates through three cascaded modules:
(i) \textbf{Hypothesis-Driven Decomposition}:
Acting as the "Planner," this module utilizes the LLM's semantic priors to decompose complex queries. 
(ii) \textbf{Hierarchical Active Perception}:
Acting as the "Scout," this module utilizes a Grid-Pyramid mechanism to actively zoom in on evidence from relevant temporal windows.
(iii) \textbf{Evidence Blackboard \& Arbitration}: Acting as the "Solver," this module aggregates discrete visual observations into a structured memory buffer to support reasoning. 

By decoupling reasoning planning from visual perception, TIR-Flow delivers substantial improvements across diverse benchmarks. Notably, it propels the Qwen2.5-VL-7B to challenge proprietary giants, significantly narrowing the performance gap with Gemini 1.5 Pro on EgoSchema and approaching the level of Gemini 2.5 Pro on MVBench. 
Our approach demonstrates that a frozen VLM, when equipped with the agency to actively seek information, can outperform models trained with extensive additional data engineering and training costs.

\begin{figure}[h]
  \centering
  \includegraphics[width=\linewidth,trim=10pt 10pt 10pt 10pt, clip]{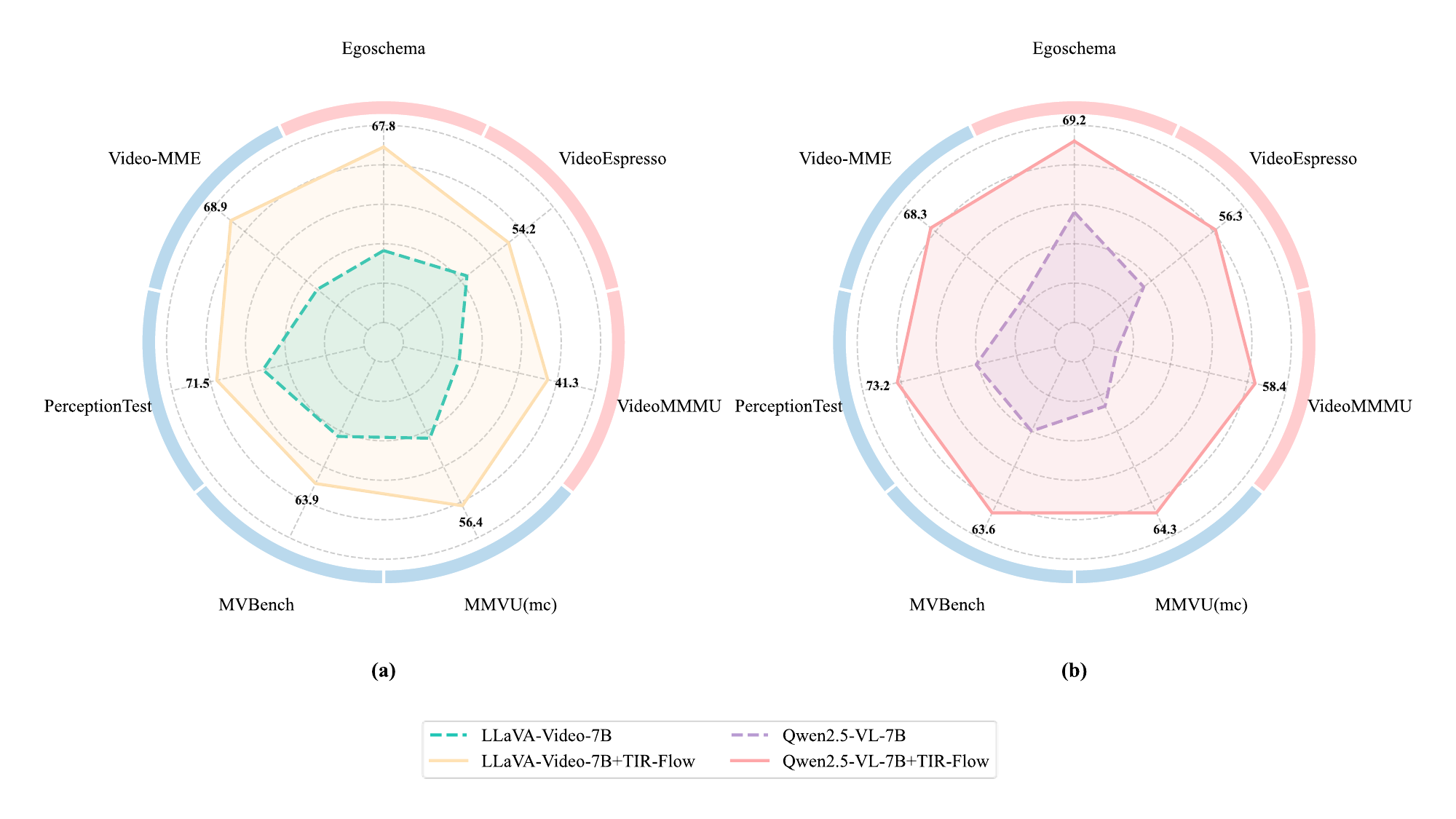}
    \caption{\textbf{Unlocking reasoning potential.} TIR-Flow consistently boosts the zero-shot performance of (a) LLaVA-Video-7B and (b) Qwen2.5-VL-7B across seven benchmarks, improving both perception and reasoning capabilities.}
  \label{fig:radar}
\end{figure}

In summary, our contributions are three-fold:
\begin{itemize}
% [leftmargin=*, topsep=2pt, itemsep=2pt, parsep=0pt, partopsep=0pt]
\item \textbf{Unlocking Reasoning Ceiling:} We propose TIR-Flow, a training-free framework shifting from static optimization (SFT/RL) to  active perception. Its System-2 ``Plan, Look, and Verify'' agency enables frozen VLMs to surpass the reasoning capabilities of supervised baselines (see~\autoref{tab:mainresults}).

\item \textbf{Bridging the Gap via Active Perception:} We introduce a closed-loop architecture synergizing decomposition, active scouting, and arbitration. This design aligns semantic planning with visual exploration, enabling the model to dynamically ``zoom in'' on critical spatiotemporal evidence to resolve information bottlenecks (see~\autoref{sec: method}).

\item \textbf{Superior Performance and Reliability:} TIR-Flow achieves best performance across seven benchmarks with a 5.9\% average boost (see~\autoref{fig:radar} and~\autoref{tab:mainresults}). Crucially, our ablation analysis confirms these gains stem from genuine evidence grounding , ensuring reliable complex reasoning (see~\autoref{sec: ablation}).
\end{itemize}

\section{Related Work}
\subsection{Video-Language Models.}

Recent years have witnessed the surge of Large Vision-Language Models (Video-LLMs) that extend static image encoders to the temporal domain~\cite{lin2024videollavalearningunitedvisual,li2025videochat,zhang2025videollama,wang2025internvl35advancingopensourcemultimodal,bai2025qwen3vltechnicalreport}.
Standard paradigms typically operate in a single pass, feed forward manner. These models passively consume pre-defined visual tokens, lacking the agency to re-examine or look back at specific regions when initial reasoning proves insufficient. Consequently, fine-grained details are either lost in compressed global embeddings or buried under an overwhelming volume of redundant tokens.

\begin{figure*}[h]
    \centering
    \includegraphics[width=1\linewidth, trim=70pt 50pt 30pt 20pt, clip]{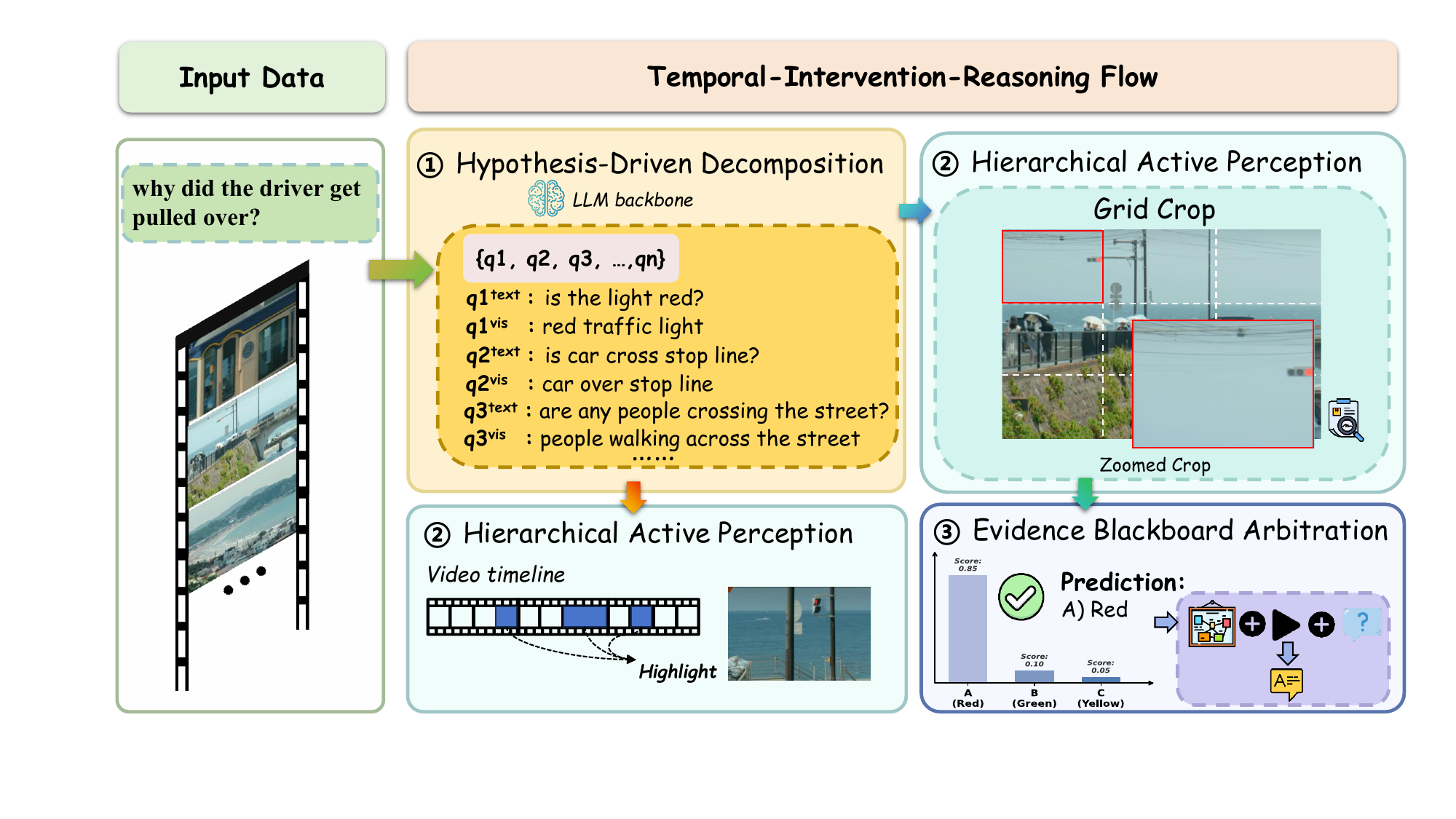}
    \caption{\textbf{Overall Architecture of the TIR-Flow.} First, the user question is decomposed by an LLM into structured sub-queries. Next, hierarchical active perception collects targeted evidence via coarse-to-fine spatial crops. Finally, an evidence blackboard arbitrates and scores the evidence, and the prediction is fused with the query to generate the final response.}
    \label{fig:overall}
\end{figure*}

\subsection{RFT for Video Reasoning.}

To enhance the reasoning capabilities of Video-LLMs, a dominant line of work pursues data-centric alignment via large-scale Video Chain-of-Thought (CoT) synthesis~\cite{wang2025videotree, li2025zebracotdatasetinterleavedvision} and post-training RL to enforce step-by-step reasoning~\cite{zhang2025videocotcomprehensivedatasetspatiotemporal, chen2025scalingrllongvideos, feng2025videor1reinforcingvideoreasoning,jin2025videomem}. However, recent studies~\cite{yue2025doesreinforcementlearningreally, huang2024largelanguagemodelsselfcorrect, kirk2024understandingeffectsrlhfllm} suggest these paradigms largely act as alignment, concentrating probability mass on pre-defined patterns rather than expanding the reasoning ceiling: pass@1 may improve, yet pass@k often stagnates or even drops. This limitation is amplified in Video-LLMs, where parameter-intensive optimization is costly and performance remains fundamentally bounded by perceptual bottlenecks in frozen encoders.

\subsection{Test Time Scaling and Active Agents}
A growing body of work studies inference-time techniques for boosting model performance~\cite{yao2023thinking,xu2025llava}, echoing System-2-style deliberation in human cognition~\cite{wang2025multimodalchainofthoughtreasoningcomprehensive,zhang2024cocot,yao2023tree,Besta2023GraphGOT}. In vision, agent-based approaches~\cite{surís2023vipergptvisualinferencepython,yang2023mmreactpromptingchatgptmultimodal,fan2024videoagent} decompose tasks into sub-steps and invoke external tools, but often depend on fragmented toolchains~\cite{wang2025mllmtoolmultimodallargelanguage, gao2025multimodalagenttuningbuilding}, leading to error propagation and maintenance overhead. Meanwhile, training-free methods such as SmartSight~\cite{sun2025smartsightmitigatinghallucinationvideollms}, VideoTree~\cite{wang2025videotree}, InfiniPot-V~\cite{kim2025infinipotvmemoryconstrainedkvcache}, and STTM~\cite{hyun2025multigranularspatiotemporaltokenmerging} mitigate the tension between redundancy and hardware constraints, but typically trade off memory, and reasoning fidelity. In contrast, TIR-Flow offers a unified endogenous framework that leverages a VLM’s native search for lightweight yet high-resolution video reasoning, enabling effective inference-time scaling.

\section{Method}\label{sec: method}
\subsection{Overview and Problem Formulation}
We address the problems of VideoQA under a training-free setting. Given a long video $\mathcal{V} = \{v_1, v_2, ..., v_T\}$ and a complex natural language query $Q$, our goal is to generate an accurate answer $A$ without fine-tuning the underlying Vision-Language Model (VLM).

Traditional approaches maximize $P(A|\mathcal{V}, Q)$ on fixed token sequences but suffer from redundancy and attention dilution. Instead, we formulate reasoning as a Modularized Flow, decomposing it into discrete steps where the model actively seeks visual evidence to verify sub queries. The inference process is modeled as:
\begin{equation}
\small
A^*=\arg\max_{A}\sum_{\mathcal{P},\,\mathcal{E}}
P(A\mid \mathcal{V},\mathcal{E},\mathcal{P})
P(\mathcal{E}\mid \mathcal{V},\mathcal{P})
P(\mathcal{P}\mid Q).
\end{equation}
where $\mathcal{P}$ represents the reasoning plan,
$\mathcal{V}$ represents the video features and $\mathcal{E}$ represents the retrieved visual evidence. 
To implement this, we propose Temporal-Intervention-Reasoning Flow.

In~\autoref{fig:overall},  we illustrate the framework of TIR-Flow, which enhances VLMs by following rules to fetch the right visual evidence during inference. 

\subsection{Hypothesis-Driven Decomposition}\label{sec:HDD}
Complex queries (e.g., "Why did the driver get pulled over?") often lack direct visual mappings. The HDD module serves as a cognitive "Planner." It utilizes the semantic reasoning capability of the LLM backbone to decompose the user query $Q$ into a structured reasoning tree $\mathcal{T} = \{q_1, q_2, ..., q_N\}$.

At step $i$, the planner generates $(q_i^{text}, q_i^{vis})$, representing the logical query and a visual matching description, respectively.
\begin{equation}
    \mathcal{P} = \text{LLM}_{plan}(Q) \rightarrow \{(q_1^{text}, q_1^{vis}), ...\}
\end{equation}
This decomposition transforms abstract causal inquiries into verifiable visual existence checks.

\subsection{Hierarchical Active Perception}\label{sec:HAP}

\begin{figure*}[t]
    \centering
    \includegraphics[width=\linewidth, trim=30pt 110pt 10pt 10pt, clip]{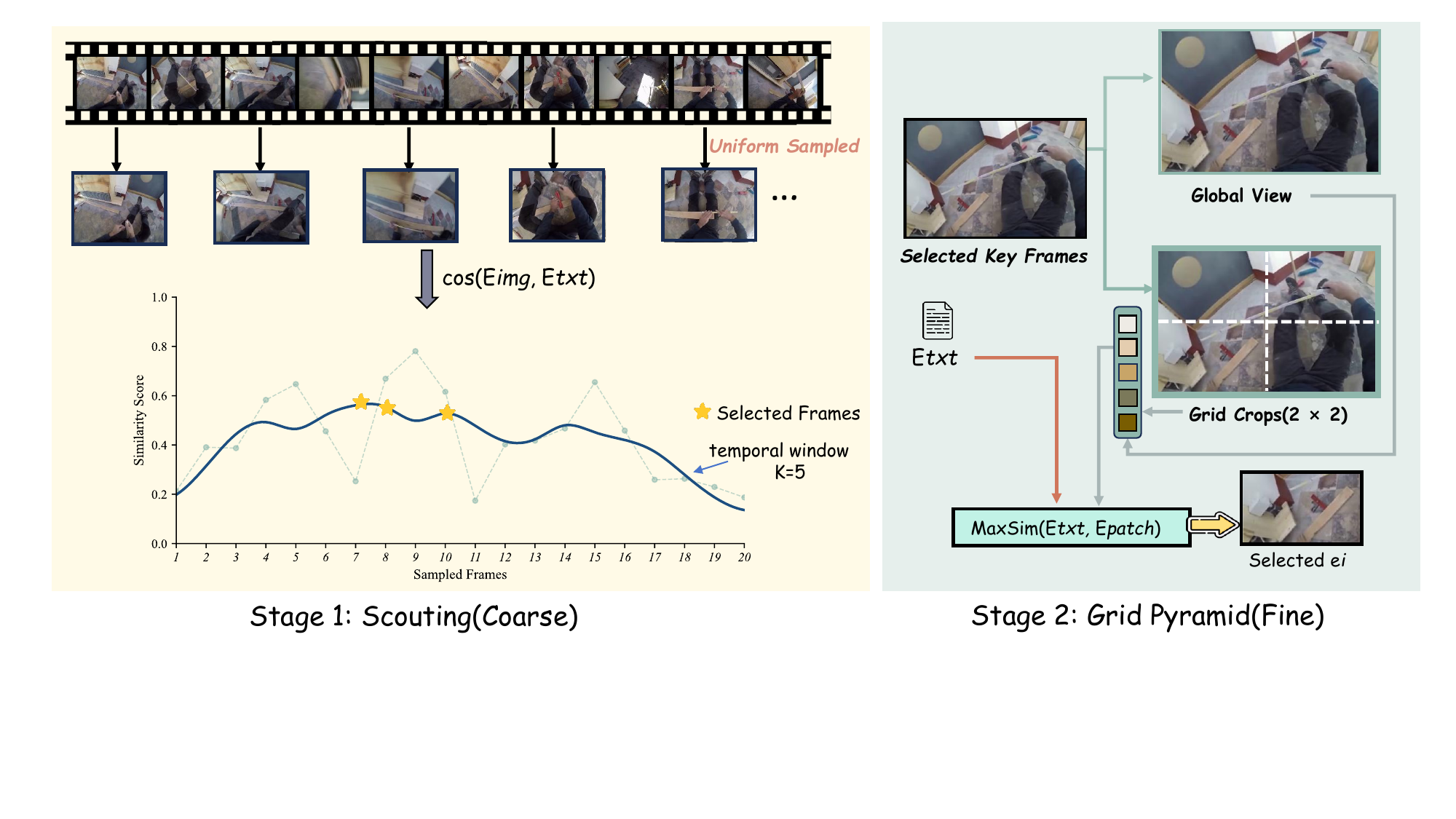}
    \caption{\textbf{Two Stage Grid Crop Pipeline:} Similarity Based Keyframe Scouting ($k=5$) Followed by Grid-Pyramid Patch Retrieval.}
    \label{fig:gridcrop}
\end{figure*}

To efficiently locate visual evidence $\mathcal{E}$ without processing the entire high-resolution video, we design a coarse-to-fine retrieval mechanism involving Temporal Scouting and Spatial Focusing.

\noindent \textbf{Temporal Scouting.}
We first identify the relevant temporal windows. We employ a lightweight, dual-encoder model (e.g., CLIP) to extract frame-level features $F_t$ from the video at a fixed sampling rate. For a visual query $q_i^{vis}$, we compute the cosine similarity scores
\begin{equation}
    S_t = \cos(E_{img}(v_t), E_{txt}(q_i^{vis}))
\end{equation}
Instead of selecting discrete frames, we apply a temporal smoothing window (kernel size $k=5$) to capture continuous actions and select the top-$K$ time intervals $\mathcal{W} = \{w_1, ..., w_K\}$ with the highest aggregated scores.

\noindent \textbf{Grid Pyramid Strategy.}
Standard resizing of images often obliterates small but crucial details (e.g., a traffic light in the distance). To address this, we introduce a Spatial Grid-Pyramid strategy.
For a candidate keyframe $I \in \mathcal{W}$, we construct a set of patches $\mathbb{P}$ containing:
(i) The Global View: $I_{global}$ (resized to preserve context).
(ii) Local Grid Crops: We divide $I$ into an $N \times N$ grid. 

We then employ the CLIP visual encoder to score these patches against $q_i^{vis}$. The most relevant spatial region is selected as the visual evidence $e_i$:
\begin{equation}
    e_i = \underset{p \in \mathbb{P}}{\arg\max} \left( \text{Sim}(E_{img}(p), E_{txt}(q_i^{vis})) \right)
\end{equation}
This mechanism allows the model to zoom in on relevant regions dynamically, effectively increasing the effective resolution by a factor of $N^2$.

\subsection{Evidence Blackboard \& Arbitration}\label{sec: EBA}
Unlike traditional approaches that treat reasoning as a linear chain, we model the evidence accumulation as a dynamic state maintenance process. The EBA module serves as the central memory unit, maintaining a Evidence Blackboard $\mathcal{B}_t$, which stores a structured history of verified visual facts in natural language form.

Evidence Arbitration Mechanism.
At each reasoning step $t$, the VLM acts as an "Arbitrator" to validate the newly retrieved visual evidence $e_t$ (from~\autoref{sec:HAP}) against the current blackboard state $\mathcal{B}_{t-1}$ and the sub-query $q_t^{text}$. Formally, we define the arbitration function as:
\begin{equation}
  (\mathcal{O}_t, s_t, \gamma_t) = \text{VLM}_{\text{arb}}(e_t, q_t^{text}, \mathcal{B}_{t-1})  
\end{equation}
Here, $\mathcal{O}_t$ is the textual observation from $e_t$, $s_t \in [0,1]$ is the confidence score, and $\gamma_t \in \{0, 1\}$ denotes a conflict indicator. We prompt the VLM to set $\gamma_t=1$ if $e_t$ contradicts $\mathcal{B}_{t-1}$ or appears ambiguous (see~\autoref{sec: prompt details} for prompt details).

State Update and Feedback Loop. Based on the arbitration output, the system executes one of two atomic operations:
(i) Update (Acceptance): If the confidence $s_t$ exceeds a threshold $\tau$ and no conflict is detected ($\gamma_t=0$), the observation is accepted. The blackboard state updates by appending the new fact: $\mathcal{B}_t = \mathcal{B}_{t-1} \cup \{\mathcal{O}_t\}$.
(ii) Refine (Rejection): If $s_t < \tau$ or $\gamma_t=1$, the evidence is deemed unreliable. The EBA triggers a Refinement Signal back to the HDD module (see~\autoref{sec:HDD}).
This signal carries the error type (e.g., "object occlusion" or "temporal mismatch"), prompting the HDD to regenerate a more granular sub-query $\hat{q}_t$ to re-examine the video.

This distinct separate-then-verify mechanism ensures that the final reasoning is grounded solely on high-confidence, non-contradictory visual evidence, significantly reducing hallucinations.

\begin{table*}[t]
\centering
\resizebox{\textwidth}{!}{%
\begin{tabular}{lcccccccc} % 9列：Models + Frames + 7个指标
\toprule

\multirow{2}{*}{\textbf{Models}}
& \multirow{2}{*}{\textbf{Frames}}  % 新增Frames列（居中）
& \multicolumn{4}{c}{\textbf{Reasoning}}
& \multicolumn{3}{c}{\textbf{Perception}}
\\
\cmidrule(lr){3-6}\cmidrule(lr){7-9}
& & Egoschema & VideoEspresso & VideoMMMU & MMVU(mc) & MVBench & PerceptionTest & VideoMME
\\

\midrule
\multicolumn{9}{l}{\textit{Proprietary models}} \\
\rowcolor{gray!8}
GPT-4o & 64 & - & - & 61.2 & \textbf{75.4} & - & - & - \\
\rowcolor{gray!8}
Gemini-1.5-Flash & 128 & 65.7 & 39.8 & 49.8 & - & - & - & 76.1 \\
\rowcolor{gray!8}
Gemini 1.5 Pro & 128 & \textbf{72.2} & 44.2 & 53.9 & - & - & - & \underline{78.6} \\
\rowcolor{gray!8}
Gemini 2.5 Pro & - & - & \textbf{83.6}  & \textbf{74.9}  & - & \textbf{69.9} & - & \textbf{85.1}  \\

\midrule
\multicolumn{9}{l}{\textit{Training-Free models}} \\
SmartSight$^{\dagger}$ & 32 & - & - & 47.6 & - & - & - & 56.2 \\
% CogniGPT$^{\dagger}$ & < 20 & 69.2 & - & - & - & - & - & 54.7 \\
VideoTree$^{\dagger}$ & - & 66.2 & - & - & - & - & - & 53.1 \\
InfiniPot-V$^{\dagger}$ & 768 & 65.8 & - & - & - & - & - & 61.1 \\
STTM$^{\dagger}$ & 1fps & 58.6 & - & - & - & - & - & 62.6 \\

\midrule
\multicolumn{9}{l}{\textit{Training models}} \\
LLaVA-Mini-8B & 1fps & 51.2 & - & - & - & 44.5 & - & - \\
VILA-40B & 256 & 58.0 & - & 34.0 & - & - & 54.0 & - \\
LLaVA-onevision-7B &32& 60.1 & 56.1* & 57.1 & 49.2 & 33.9 \\
VideoLLaMA3-7B & 180 & 63.3 & - & - & 44.1 & \underline{69.7} & \underline{72.8} & 66.2 \\
Video-R1-7B & 32 & - & - & 52.3 & 64.2 & 63.9 & - & 59.3 \\

Owen3-VL-8B-Instruct & 2fps & 60.6* & 52.0* & \underline{65.3} & \underline{65.6} & 68.7  & 63.7* & 71.4 \\
% Qwen3-VL-8B-Thinking & 2fps & 41.4* & - & 72.8 & (62.0) & 69.0 & - & 71.8 \\
\midrule

LLaVA-Video-7B & 32 & 57.3 & 48.8 &34.4 & 48.8 & 58.6&67.9 &63.3 \\
\rowcolor{gray!20}
\textbf{LLaVA-Video-7B+TIR-Flow} & 32 &67.8~\textcolor{red}{(10.5$\uparrow$)}&54.2~\textcolor{red}{(5.4$\uparrow$)}&41.3~\textcolor{red}{(6.9$\uparrow$)}&56.4~\textcolor{red}{(7.6$\uparrow$)}& 63.9~\textcolor{red}{(5.3$\uparrow$)}&71.5~\textcolor{red}{(3.6$\uparrow$)}&68.9~\textcolor{red}{(5.6$\uparrow$)} \\
Qwen2.5-VL-7B & 32&65.6&47.0&51.2& 61.3 &59.0&69.1&62.4\\
\rowcolor{gray!20}
\textbf{Qwen2.5-VL-7B+TIR-Flow} & 32 &\underline{69.2}~\textcolor{red}{(3.6$\uparrow$)} &\underline{56.3}~\textcolor{red}{(9.3$\uparrow$)} & 58.4~\textcolor{red}{(7.2$\uparrow$)} &64.3~\textcolor{red}{(3.0$\uparrow$)} &63.6~\textcolor{red}{(4.6$\uparrow$)} &\textbf{73.2}~\textcolor{red}{(4.1$\uparrow$)}&68.3~\textcolor{red}{(5.9$\uparrow$)} \\
\bottomrule
\end{tabular}%
}
\caption{Performance on reasoning and perception video QA multiple choice benchmarks. * indicates the result we reproduced using a 32-frame setting. $^{\dagger}$ indicates training-free framework. The best performance for each metric is \textbf{bolded}, and the second best is \underline{underlined}.}
\label{tab:mainresults}
\end{table*}

\section{Experiment}
\subsection{Settings}

\noindent \textbf{Baselines.} We evaluate TIR-Flow against a diverse set of state-of-the-art methods, categorized into:
(i) Proprietary Models: GPT-4o~\cite{openai_gpt4o_2024}, Gemini~\cite{comanici2025gemini}.
(ii) Open-source Models: VideoLLaMA3-7B~\cite{zhang2025videollama}, Video-R1-7B~\cite{feng2025videor1reinforcingvideoreasoning}, Qwen3-VL-8B~\cite{bai2025qwen3vltechnicalreport}, and LLaVA-OneVision-7B~\cite{li2024llavaonevisioneasyvisualtask}.
(iii) Training-Free Frameworks: We include SmartSight~\cite{sun2025smartsightmitigatinghallucinationvideollms} (built on Qwen2.5-VL), VideoTree~\cite{wang2025videotree}, InfiniPot-V~\cite{kim2025infinipotvmemoryconstrainedkvcache}, and STTM~\cite{hyun2025multigranularspatiotemporaltokenmerging}.
Please refer to the~\autoref{sec:baseline} for more details.

\noindent \textbf{Implementation Details.}
We separately use Qwen2.5-VL-7B and LLaVA-video-7B as our base models. 
We represent each video as a sequence with maximum $T$ frames. 
$T$ is initialized to 32 here. Regarding the hyperparameters, we set the window size $k=5$, the Top-$K$ selection to 3, and the EBA judgment threshold $\tau=0.7$. The input frame $I$ is spatially partitioned into an $N \times N$ grid (with $N=3$). All experiments are conducted on NVIDIA V100 (32GB) GPUs.

\subsection{Results}
In this section, 
We evaluate TIR-Flow against Qwen2.5-VL and LLaVA-Video across reasoning  and perception benchmarks. As shown in~\autoref{tab:mainresults}, our training-free framework demonstrates competitive performance against state-of-the-art baselines, effectively bridging the gap between raw visual input and complex reasoning.

\noindent \textbf{TIR-Flow significantly improves performance on reasoning-intensive tasks.} Experimental results in~\autoref{tab:mainresults} demonstrate our framework's effectiveness. While proprietary models (e.g., Gemini 1.5 Pro) achieve strong results via costly alignment, and training-free methods like VideoTree and STTM offer only marginal gains, TIR-Flow delivers substantial improvements without parameter updates. For instance, TIR-Flow boosts average accuracy by \textbf{6.7\%} on multiple reasoning benchmarks. With this framework, Qwen2.5-VL-7B attains 69.2\% on EgoSchema and consistent gains of 7.2\% and 9.3\% on VideoMMMU and VideoEspresso, respectively. These findings suggest that task-guided active search unlocks intrinsic reasoning potential of frozen VLMs, transforming latent knowledge into structured deductions.

We attribute this to Perception-Reasoning Decoupling. Traditional models often succumb to the vanishing detail problem by entangling feature extraction and deduction. By offloading evidence search to HAP and EBA modules, TIR-Flow frees the LLM backbone to function purely as a cognitive orchestrator. This implies the primary bottleneck in VLMs is not reasoning capacity, but the high entropy of the input space, which TIR-Flow manages through structured deductions.

\noindent \textbf{TIR-Flow notably enhances perception-heavy task performance.} As validated in~\autoref{tab:mainresults}, our framework effectively mitigates perceptual limitations caused by information dilution in traditional static sampling. TIR-Flow addresses this bottleneck via a hierarchical zoom-in strategy for targeted feature acquisition. This mechanism ensures that critical visual semantics are preserved with high fidelity. While base models like LLaVA-Video often fail to resolve subtle details, LLaVA-Video-7B with TIR-Flow achieves a \textbf{5.6\%} gain on VideoMME. This leap reflects a Signal-to-Noise Ratio (SNR) advantage: active search acts as a dynamic filter, selectively boosting salient visual patches while suppressing noise. By relying on the model's intrinsic semantic understanding rather than memorized data patterns, our approach maintains robustness even when encountering novel scenes or complex visual compositions unseen during pre-training.

In summary, these results demonstrate that TIR-Flow offers a significant improvement in both reasoning and perception tasks by effectively shifting the computational burden from parameter optimization to active perception.

\section{Ablation Studies}\label{sec: ablation}
\subsection{Effectiveness of different modules}
To further reveal the underlying mechanism of complex video reasoning, we conduct an ablation study by evaluating three variants of TIR-Flow, each distinct in its component configuration:
(i) TIR-Flow w/o HDD: Bypasses decomposition, directly utilizing the original query $Q$ as the probe for HAP-based evidence retrieval.
(ii) TIR-Flow w/o HAP: Eliminates the active spatiotemporal scouting mechanism. It directly feeds uniformly sampled frames alongside the sub queries into the EBA module for evidence extraction.
(iii) TIR-Flow w/o EBA: Omits the filtering and arbitration mechanisms, directly accumulating all retrieved evidence for final inference without validity verification.

\begin{table}[h] 
\scriptsize 
\centering
\setlength{\tabcolsep}{6pt} 
\begin{tabular}{lcccc}
\toprule[1.pt]
\textbf{Modules} & \textbf{Ego} & \textbf{Espresso} & \textbf{MVBench} & \textbf{Per}\\
\midrule
Base               & 57.3/65.5 & 48.8/47.0 & 58.6/59.0 & 67.9/69.1\\
+HAP+EBA      & +5.8/+0.5 & +2.2/+4.5 & +1.8/+0.6 & +0.2/+0.3\\
+HDD+HAP &  +8.0/+2.1 & +4.5/+7.2 & +3.5/+2.9 & +2.6/+2.7\\
+HDD+EBA          & +2.6/+1.2 & +2.9/+2.5 & +2.3/+2.2 & +2.4/+1.9\\
\rowcolor{gray!20}
+HDD+HAP+EBA      & +10.5/+3.7 & +5.4/+9.3 & +5.3/+4.6 & +3.6/+4.1\\
\bottomrule[1.pt]
\end{tabular}
\caption{\textbf{Ablation Study.} Base model is LLaVA/Qwen. 
\textbf{Ego}: Egoschema.
\textbf{Per}: PerceptionTest.}
\label{tab:effectiveness_single}
\end{table}

As shown in~\autoref{tab:effectiveness_single}, the full TIR-Flow framework achieves the best performance across all benchmarks, validating the synergy of the proposed modules. We analyze the contribution of each component based on the quantitative drops observed when they are removed:

\textbf{Impact of Hierarchical Decomposition.} The HDD module proves indispensable for reasoning-intensive tasks, serving as the cognitive roadmap for complex queries. When removing HDD, we observe the sharpest performance decline on EgoSchema, dropping from 10.5\% to 5.8\%. This empirical evidence confirms that without structured task decomposition, the model struggles to bootstrap the complex logical deductions required for long-form video reasoning, effectively reverting to shallower understanding.

\textbf{Impact of Active Perception.} The HAP module is critical for handling fine visual details that are typically lost in standard downsampling processes. The removal of HAP leads to a noticeable drop on perception benchmarks like MVBench (from 5.3\% down to 3.3\%) and EgoSchema (decreasing by 3.3\%). This validates that HAP effectively addresses the perceptual resolution bottleneck. By employing an active strategy, it increases the visual signal-to-noise ratio, ensuring subsequent reasoning is based on high-fidelity evidence.

\textbf{Impact of Evidence Arbitration.} Finally, the exclusion of EBA results in consistent performance degradation, exemplified by a 2.5\% drop on EgoSchema and a 1.8\% drop on MVBench. This widespread decline underscores the critical importance of a structured memory buffer for cross verifying disparate evidence. Video streams inherently contain redundant and occasionally ambiguous information; without EBA's filtering and arbitration mechanisms, the model accumulates noisy or logically contradictory observations. This error propagation diminishes the model's capacity to resolve visual ambiguities, leading to lower confidence and accuracy during the final inference stage.

\subsection{Micro-Ablation}
Spatiotemporal Synergy within HAP.
The HAP module is built upon two pillars: \textit{Temporal Scouting} for identifying critical moments and \textit{Spatial Focusing} (Grid Pyramid) for resolving fine-grained details. To disentangle their individual contributions, we evaluate two sub-variants on representative benchmarks: EgoSchema and MVBench.
\textbf{w/o Spatial Focusing:} This variant retains the temporal scouting mechanism to retrieve relevant time windows but disables the Grid Pyramid strategy. The selected frames are processed solely via the global view (standard resizing), removing the model's ability to zoom in. \textbf{w/o Temporal Scouting:} This variant replaces the relevance based temporal scouting with standard uniform sampling. While it retains the Grid Pyramid for spatial enhancement, the model blindly distributes its high-resolution budget across the video, potentially missing sparse temporal cues.

\begin{table}[h]
\scriptsize
\centering
\setlength{\tabcolsep}{6pt}
\begin{tabular}{l|cc|cc}
\toprule
\multirow{2}{*}{\textbf{Configuration}} & \multicolumn{2}{c|}{\textbf{Mechanism}} & \multirow{2}{*}{\textbf{EgoSchema}} & \multirow{2}{*}{\textbf{VideoMME}} \\
& Temporal & Spatial & & \\
\midrule
w/o Spatial & \checkmark & $\times$ &65.2/67.5 &67.0/66.9 \\
w/o Temporal & $\times$ & \checkmark &63.4/65.6&66.8/66.5 \\
w/o S\&T& $\times$ &$\times$ &59.9/66.7 &64.2/63.1 \\
\midrule
\textbf{Full HAP} & \checkmark & \checkmark & \textbf{67.8/69.2} & \textbf{68.9/68.3} \\
\bottomrule
\end{tabular}
\caption{\textbf{Component Analysis within HAP.} Base model is LLaVA/Qwen. We report accuracy (\%) on EgoSchema and VideoMME.}
\label{tab:hap_micro_ablation}
\end{table}

\noindent \textbf{Analysis.} Table~\ref{tab:hap_micro_ablation} reveals the distinct roles of each component:
Temporal Scouting dominates long-horizon reasoning. On EgoSchema benchmark, which relies on locating sparse cues in long videos, removing the temporal mechanism causes a more severe drop (69.2\% $\to$ 65.6\%) compared to removing spatial focusing (69.2\% $\to$ 67.5\%). This confirms that blindly applying high resolution to the wrong timeframe is futile, locating the \textit{critical window} is the prerequisite for success.
Spatial Focusing resolves information bottlenecks. On VideoMME dataset, even when the correct temporal window is identified, the performance drops from 68.9\% to 67.0\% without spatial focusing. This indicates that standard resolution even at the right moment fails to capture the fine-grained details required for precise answering. The best performance is achieved only when the model looks at the \textit{right time} with the \textit{right resolution}.

\subsection{Sensitivity Analysis of Hyperparameters.}
To balance model performance with computational efficiency, we conducted a comprehensive analysis of the key hyperparameters in TIR-Flow: the Sliding Window Size $k$. Additionally, we evaluate the sensitivity of the confidence threshold $\tau$ in the EBA module, which is detailed in~\autoref{sec:analysis}.

\begin{figure}[h]
  \centering
  \includegraphics[width=\linewidth, trim=0pt 320pt 500pt 0pt, clip]{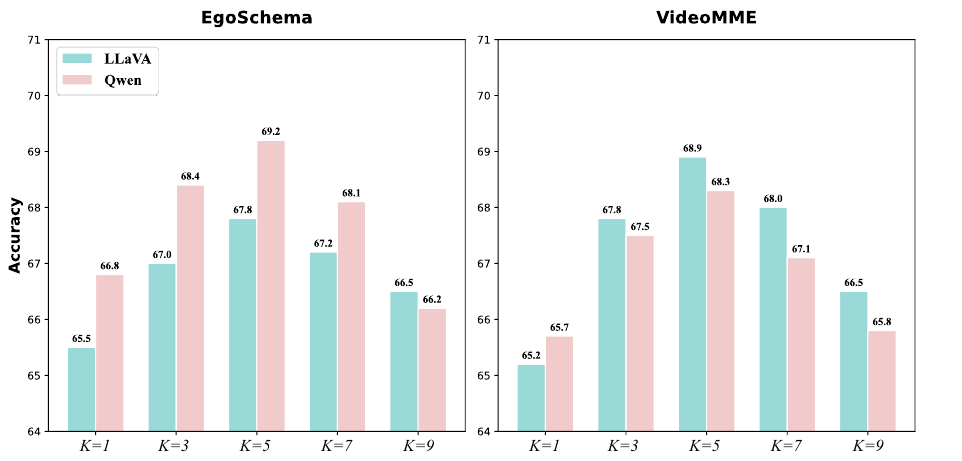}
  \caption{\textbf{Impact of Diverse Sample Size ($k$)} on EgoSchema and VideoMME.}
  \label{fig: N_influence}
\end{figure}

\noindent \textbf{Effectiveness of Sliding Window Size $k$.}
To investigate the impact of temporal granularity on active perception, we evaluate the performance of the HAP module across various sliding window sizes, specifically $k \in \{1, 3, 5, 7, 9\}$. As illustrated in~\autoref{fig: N_influence}, we observe that performance follows a bell shaped trend as $k$ increases. The exclusion of the sliding window (equivalent to $k=1$) leads to an average degradation of 2.75\%, confirming that localized temporal anchoring is essential for mitigating interference from redundant frames.
Notably, $k=5$ emerges as the optimal configuration.
Smaller window sizes ($k \le 3$) tend to oversegment the video,
potentially fracturing the continuity of complex actions, while excessively large windows ($k \ge 9$) suffer from information dilution, where the model's attention is insufficiently focused on salient events. These results suggest that an appropriately sized sliding window acts as a crucial temporal filter, allowing TIR-Flow to strike an optimal balance between local visual precision and global temporal context.

\subsection{Oracle-Based Evidence Quality Assessment}
To rigorously validate the information density of the retrieved evidence, we conduct a cross-model evaluation using \textbf{Qwen3-VL-32B-Instruct} as an external "Oracle Judge." 

\paragraph{Settings.} We randomly sample 200 instances from VideoMME. For each instance, we compare the \textbf{Average Relevance Score (ARS)} of the visual inputs provided by two methods:
\textbf{Baseline (Uniform Sampling):} We uniformly sample $N=16$ frames from the video. To quantify the signal-to-noise ratio, we feed these frames individually to the Oracle Judge. The final score for the baseline is calculated as the mean of the scores of all 16 frames.
\textbf{Ours (Active HAP):} We take the single, final zoomed-in image crop ($e_i$) selected by the HAP module. This frame is scored once by the Oracle.

The Oracle Judge is prompted to rate each image on a scale of 1-5 based on whether it provides critical visual evidence to answer the user's query.
Prompt details are shown in~\autoref{sec: prompt details}.
\begin{table}[h]
    \centering
     \resizebox{\linewidth}{!}{
    \begin{tabular}{l|cc}
    \toprule
    \textbf{Method} & \textbf{Avg. Score (1-5)} & \textbf{High-Sufficiency Rate} \\
    \midrule
    Uniform Sampling & 2.88 & 39.5\% \\
    \rowcolor{gray!10} \textbf{HAP (Ours)} & \textbf{4.15} & \textbf{71.0\%} \\
    \bottomrule
    \end{tabular}
    }
    \caption{\textbf{Oracle Evaluation of Visual Evidence Quality.} We compare the sufficiency of evidence provided by 16 uniform frames versus a single HAP-selected crop. The "High-Sufficiency Rate" denotes the percentage of samples receiving a score $\ge 4$.}
    \label{tab:oracle_eval}
\end{table}

The results in ~\autoref{tab:oracle_eval} reveal a substantial gap in evidence quality between passive sampling and our active approach. (1) Superior Information Density: While the uniform baseline processes 16 frames to achieve a modest average score of 2.88, our HAP module achieves a significantly higher score of \textbf{4.15} using only a single, strategically selected crop. This indicates that HAP effectively filters out temporal redundancy and spatial noise, concentrating the visual signal into a highly informative region. (2) Decisive Evidence Retrieval: The High-Sufficiency Rate serves as a proxy for answering confidence. The dramatic improvement from 39.5\% to \textbf{71.0\%} demonstrates that standard downsampling often misses fine-grained details required for complex queries. In contrast, TIR-Flow's active zooming mechanism successfully locates and verifies decisive visual clues that are otherwise lost, validating the necessity of active perception.

\section{Conclusion}
In this work, we have introduced TIR-Flow, a new, training-free framework for unlocking the reasoning potential of VLMs in video understanding by reformulating VideoQA as an iterative hypothesis-testing process. By integrating our plug-and-play HDD module which translates complex queries into structured plans with the high-fidelity HAP module that "zooms in" on fine-grained details and the EBA module for robust evidence arbitration, our approach effectively overcomes information redundancy and the perceptual bottleneck. Experimental results across multiple benchmarks demonstrate that TIR-Flow achieves substantial performance gains, outperforming state-of-the-art zero-shot baselines and rivaling fully supervised methods while maintaining strict logical consistency. By offering a scalable, inference-time active search solution, TIR-Flow provides a practical and explainable path forward for the next generation of intelligent video understanding systems.

\section*{Limitations}
This study has a few limitations. Firstly, due to resource constraints, our evaluation focuses on 7B-parameter models (LLaVA-Video and Qwen2.5-VL); investigating scalability on larger foundation models remains future work. Secondly, the framework relies on the base model's planning capability, where flawed initial decomposition (HDD) may propagate errors. Finally, while TIR-Flow significantly elevates reasoning ceilings, the iterative nature of System-2 active perception entails a modest trade-off in inference speed compared to static, single-pass baselines.

\bibliographystyle{acl_natbib} 
\bibliography{main}

\clearpage
\appendix
\section{Failure Case Analysis}
\label{sec:appendix-limitation}

While TIR-Flow significantly elevates the reasoning ceiling of Video-LLMs by introducing an iterative "Plan-Look-Verify" mechanism, we observe performance boundaries on tasks that rely heavily on atomic visual perception rather than logical deduction. In this section, we analyze two representative failure cases from the VideoMME benchmark to delineate the distinction between reasoning failures and perception bottlenecks.

\begin{tcolorbox}[
    colframe=black,          % 边框颜色
    colback=gray!10,         % 背景颜色
    sharp corners=southwest, % 左下角尖角
    enhanced,
    breakable,               % 自动分页和换行
    listing only,            % 使用 listing 显示代码内容
    listing options={        % listing 的设置
        basicstyle=\ttfamily\small, % 字体样式
        breaklines=true,            % 自动换行
    }
]
\textbf{Dataset:} \textit{VideoMME}

\textbf{Video:} \textit{24i4ncHuf6A}

\textbf{Question:} \textit{According to the video, how many individuals were in the car when Archduke Franz Ferdinand was assassinated?}

\textbf{Answer:} \textit{A. Three}

\textbf{Candidates:}
\begin{itemize}[noitemsep,topsep=0pt,parsep=0pt,partopsep=0pt]
    \item \textit{A. Three}
    \item \textit{B. Two}
    \item \textit{C. One}
    \item \textit{D. Four}
\end{itemize}
\end{tcolorbox}

Analysis: 
TIR-Flow successfully locates the correct temporal segment (Temporal Scouting).
However, counting is a fundamental System-1 capability.
Even with the correct keyframe and a zoomed-in view, the base VLM struggles with dense object detection and occlusion reasoning .

Insight: TIR-Flow acts as a cognitive amplifier; it can guide the model where to look, but it cannot rectify the model's intrinsic inability to recognize or count objects if the visual encoder's resolution or the LLM's numeracy is insufficient. The "Planner" cannot decompose "Count objects" into simpler text-based sub-steps effectively.

\begin{tcolorbox}[
    colframe=black,          % 边框颜色
    colback=gray!10,         % 背景颜色
    sharp corners=southwest, % 左下角尖角
    enhanced,
    breakable,               % 自动分页和换行
    listing only,            % 使用 listing 显示代码内容
    listing options={        % listing 的设置
        basicstyle=\ttfamily\small, % 字体样式
        breaklines=true,            % 自动换行
    }
]
\textbf{Dataset:} \textit{VideoMME}

\textbf{Video:} \textit{LCtOpCi5r2s}

\textbf{Question:} \textit{Which item was not featured in the video?}

\textbf{Answer:} \textit{A. Three}

\textbf{Candidates:}

\begin{itemize}[noitemsep,topsep=0pt,parsep=0pt,partopsep=0pt]
  \item \textit{A. Balance scale}
    \item \textit{B. Traffic light}
    \item \textit{C. Gavel}
    \item \textit{D. Magnifying glass}
\end{itemize}
\end{tcolorbox}

Analysis:
Unlike causal questions (e.g., "Why did the car stop?"), which have strong semantic cues to guide the Grid-Pyramid search (e.g., look for traffic lights), existence verification requires scanning the entire video for all candidates.
The "Planner" attempts to verify the existence of each item. However, if an object (e.g., a magnifying glass) appears briefly in a cluttered background, the VLM's lack of fine-grained discriminative power leads to false negatives. 

Insight: TIR-Flow is optimized for hypothesis-driven verification (confirming a specific clue exists), not exhaustive open-set scanning. When the semantic link between the query and the visual evidence is weak, the active search mechanism degenerates to the base model's passive scan, failing to compensate for the "blind spots" in low-level perception.

\section{Prompt Engineering Details}
\label{sec: prompt details}

\definecolor{promptgray}{RGB}{245,245,245}
\definecolor{promptborder}{RGB}{180,180,180}

\lstdefinestyle{promptstyle}{
  basicstyle=\ttfamily\small,
  backgroundcolor=\color{promptgray},
  frame=single,
  rulecolor=\color{promptborder},
  breaklines=true,
  breakatwhitespace=true,
  columns=fullflexible,
  keepspaces=true
}

\newtcolorbox{promptbox}[2][]{
  colback=promptgray,
  colframe=promptborder,
  fonttitle=\bfseries,
  title=#2,
  boxrule=0.8pt,
  arc=2pt,
  left=6pt,
  right=6pt,
  top=6pt,
  bottom=6pt,
  #1
}

\begin{promptbox}{HDD Planner Prompt:
Hypothesis Decomposition}
\begin{lstlisting}[style=promptstyle]
You are a video reasoning planner.

Given a complex video question, your task is to decompose it into a set of independent, visually verifiable sub-questions.

For each sub-question:
1. Provide a concise textual question (q_text).
2. Provide a short visual query phrase (q_vis) suitable for visual matching.

Guidelines:
- Each sub-question should test a single hypothesis.
- q_vis must refer to observable entities or actions.
- Avoid abstract concepts such as intentions or motivations.
- Do NOT answer the questions.

Output format (JSON list):
[
  {"q_text": "...", "q_vis": "..."},
  ...
]

Original Question:
<QUESTION>
\end{lstlisting}
\end{promptbox}

\begin{promptbox}{HDD Refinement Prompt}
\begin{lstlisting}[style=promptstyle]
HDD Refinement Prompt:
You are a strategic planner refining a failed visual search.
The previous search for the sub-question failed.
Input:
1. Original sub-question:
   q_text: <q_text>
   q_vis: <q_vis>
2. Error Diagnosis (from Arbitrator):
   <error_type> (e.g., "Object Occlusion", "Temporal Mismatch", "Low Confidence", "Contradictory Evidence")

Your task is to regenerate a more granular sub-query to overcome this specific error.

Guidelines:
- If Error is "Object Occlusion/Blur": Switch to looking for associated static objects or surrounding context.
- If Error is "Temporal Mismatch": Look for temporal landmarks (e.g., "before X happens").
- If Error is "Contradiction": Break the question into two smaller verification steps.
- q_vis must be concrete and observable.

Output format (JSON):
{"q_text": "...", "q_vis": "..."}
\end{lstlisting}
\end{promptbox}

\begin{promptbox}{EBA Prompt: Evidence Arbitration}
\begin{lstlisting}[style=promptstyle]
You are an evidence arbitrator for video reasoning.

You are given:
1. A visual observation from the video.
2. A question about the observation.
3. A list of previously verified facts (Evidence Blackboard).

Your task:
- Decide whether the visual evidence supports the question.
- Produce a concise factual observation.
- Assign a confidence score between 0 and 1.

Rules:
- Base your judgment ONLY on the visual evidence.
- If the evidence is ambiguous, lower the confidence.
- If the observation contradicts existing evidence, mark a conflict.

Evidence Blackboard:
<BLACKBOARD>

Question:
<q_text>

Output format (JSON):
{
  "observation": "...",
  "confidence": 0.xx,
  "conflict": true/false
}
\end{lstlisting}
\end{promptbox}

\begin{promptbox}{Final Prompt: Answer Synthesis}
\begin{lstlisting}[style=promptstyle]
You are a video question answering assistant.

You are given:
1. The video frames.
2. The original question.
3. A set of verified visual facts extracted from the video.

Use the verified facts to answer the question.

Video Frames:
<VIDEO>

Original Question:
<QUESTION>

Verified Evidence:
<BLACKBOARD>

Only select the best option.
\end{lstlisting}
\end{promptbox}

\begin{promptbox}{Oracle Evidence Quality Judge}
\textbf{System Instruction:} \\
You are an expert visual evidence evaluator. Your task is to assess the quality and relevance of a specific image in the context of answering a user query about a video. You must strictly follow the scoring criteria provided.

\par\noindent\rule{\textwidth}{0.4pt} % 分割线

\textbf{User Input Template:} \\
I will provide you with a Query and a Single Image Frame extracted from a video. Please evaluate the \textbf{Information Sufficiency} of this image for answering the query.

\textbf{Query:} \texttt{\{q\_txt\}} \\
\textbf{Image Context:} \texttt{\{Context\_Note\}} \textit{(Inserted only for HAP: "Note: This image is a zoomed-in crop focusing on specific fine-grained details.")}

\textbf{Scoring Criteria (1-5):}
\begin{itemize}[leftmargin=*, nosep]
    \item \textbf{5 (Critical Evidence):} The image contains the exact visual information needed. No guessing required.
    \item \textbf{4 (Strong Evidence):} Provides strong cues. Answer can be inferred with high confidence.
    \item \textbf{3 (Partial/Related):} Relevant to the topic/scene, but specific key details are missing, blurry, or occluded.
    \item \textbf{2 (Weak Relevance):} Shows general setting but not helpful for the specific question.
    \item \textbf{1 (Irrelevant/Noise):} Completely unrelated, blurry beyond recognition, or misses the event.
\end{itemize}

\end{promptbox}

\section{Baseline Details}\label{sec:baseline}
\noindent \textbf{Baseline Setting.} We compare TIR-Flow against a wide range of state-of-the-art methods:
Proprietary Models: GPT-4o~\cite{hurst2024gpt}, Gemini-1.5-Flash~\cite{google_gemini_1.5_2024}, Gemini 1.5 Pro~\cite{google_gemini_1.5_2024} and Gemini 2.5 pro~\cite{comanici2025gemini25pushingfrontier}
Open-source Models: LLaVA-Mini-8B~\cite{zhang2025llavamini}, VILA-40B~\cite{Lin_2024_CVPR}, LLAVA-onevision-7B ~\cite{li2024llavaonevisioneasyvisualtask}, VideoLLaMA3-7B~\cite{zhang2025videollama}, Video-R1-7B~\cite{feng2025videor1reinforcingvideoreasoning}, Qwen3-VL-8B~\cite{bai2025qwen3vltechnicalreport}.
Training-Free Frameworks:  
SmartSight~\cite{sun2025smartsightmitigatinghallucinationvideollms} (built on Qwen2.5-VL-7B).
VideoTree~\cite{wang2025videotree} (built on GPT-4): A more computationally intensive baseline, utilizing an average of 62.4 frames for EgoSchema and 53.1 frames for Video-MME. InfiniPot-V~\cite{kim2025infinipotvmemoryconstrainedkvcache}  (built on LLaVA-Next-7B).  STTM~\cite{hyun2025multigranularspatiotemporaltokenmerging} (built on LLaVA-Video 7B).

\section{Data} 
To comprehensively evaluate the effectiveness of TIR-Flow across different cognitive dimensions, we adopt seven representative multiple-choice video benchmarks. We categorize these benchmarks into two groups based on their primary challenges:
Perception Benchmarks: We use MVBench~\cite{li2024mvbench}, PerceptionTest~\cite{patraucean2023perception}, and Video-MME~\cite{fu2025video}. These benchmarks demand high-fidelity spatial-temporal recognition, testing the model's ability to discern fine-grained actions, object attributes, and long-range temporal consistency.
Reasoning Benchmarks: We evaluate on EgoSchema~\cite{DBLP:conf/nips/MangalamAM23}, VideoEspresso~\cite{han2025videoespresso}, VideoMMMU~\cite{hu2025video}, and MMVU~\cite{zhao2025mmvu}. These datasets require advanced cognitive abilities beyond simple recognition, such as causal induction, intentionality analysis, and the integration of complex multi-modal context.

\section{Sensitivity Analysis}\label{sec:analysis}

\begin{table}[h]
    \centering
    \begin{tabular*}{\linewidth}{@{\extracolsep{\fill}}c c c @{}}
        \toprule
        \textbf{Threshold} & \textbf{EgoSchema} & \textbf{VideoMME} \\
            & \textit{(LLaVA / Qwen)} & \textit{(LLaVA / Qwen)} \\
        \midrule
         $\tau=0.5$ & 66.3 / 67.5 & 66.9 / 66.4 \\
         $\tau=0.6$ & 67.2 / 68.1 & 68.0 / 67.5 \\
         $\tau=0.7$ & \textbf{67.8} / \textbf{69.2} & \textbf{68.9} / \textbf{68.3} \\
         $\tau=0.8$ & 67.0 / 68.5 & 68.1 / 67.5 \\
         $\tau=0.9$ & 66.2 / 67.4 & 67.3 / 66.7 \\
        \bottomrule
    \end{tabular*}
    \caption{\textbf{Impact of Confidence Threshold $\tau$.} Analysis on EgoSchema and VideoMME.}
    \label{tab:ablation_tau}
\end{table}

\textbf{Impact of Confidence Threshold $\tau$.} We further investigate the sensitivity of the Evidence Blackboard \& Arbitration (EBA) module to the confidence threshold $\tau$. ~\autoref{tab:ablation_tau} demonstrates the accuracy curves across different benchmarks as $\tau$ varies from 0.5 to 0.9.
We observe that an overly low $\tau$ introduces noise into the reasoning chain (false positives), while an excessively high $\tau$ leads to conservative predictions and unnecessary backtracking. Empirically, the model achieves robust performance peaks across all datasets when $\tau$ is set to approximately \textbf{0.7}.

\end{document}